\documentclass[11pt,a4paper]{article}
\usepackage[hyperref]{acl2020}
\usepackage{times}
\usepackage{latexsym}
\usepackage{graphicx}

\usepackage{longtable}
\usepackage{hyperref}
\usepackage{url}
\usepackage{tabularx}
\usepackage{graphics}
\usepackage{multirow}
\usepackage{lscape}
\usepackage{tabu}
\usepackage{listings}\usepackage{caption}\usepackage{footnote}
\usepackage[plainruled,noend]{algorithm2e}
\usepackage{enumitem}
\usepackage{mathtools, nccmath}
\usepackage{amsmath}
\usepackage{amsfonts}
\usepackage{subcaption,siunitx,booktabs}

\usepackage{microtype}

\aclfinalcopy \def\aclpaperid{454}

\usepackage{xcolor}
 
\definecolor{myblue}{RGB}{11, 83, 148}
\definecolor{myred}{RGB}{153, 0, 0}

\title{MobileBERT: a Compact Task-Agnostic BERT \\ for Resource-Limited Devices}

\author{
Zhiqing Sun$^1$\thanks{This work was done when the first author was an intern at Google Brain.} ,
Hongkun Yu$^2$,
Xiaodan Song$^2$,
Renjie Liu$^2$,
Yiming Yang$^1$,
Denny Zhou$^2$\\
\\
$^1$Carnegie Mellon University \texttt{\{zhiqings, yiming\}@cs.cmu.edu}\\
$^2$Google Brain \texttt{\{hongkuny, xiaodansong, renjieliu, dennyzhou\}@google.com}\\
}

\date{}

\makeatletter\begin{document}
\maketitle
\begin{abstract}
	Natural Language Processing (NLP) has recently achieved great success by using huge pre-trained models with hundreds of millions of parameters. However, these models suffer from  heavy model sizes and high latency such that they cannot be deployed to resource-limited mobile devices. In this paper, we propose MobileBERT for compressing and accelerating the popular BERT model. Like the original BERT, MobileBERT is task-agnostic, that is, it can be generically applied to various downstream NLP tasks via simple fine-tuning. Basically, MobileBERT is a thin version of $\text{BERT}_\text{LARGE}$, while equipped with bottleneck structures and a carefully designed balance between self-attentions and feed-forward networks. To train MobileBERT, we first train a specially designed teacher model, an inverted-bottleneck incorporated $\text{BERT}_\text{LARGE}$ model. Then, we conduct knowledge transfer from this  teacher to MobileBERT. Empirical studies show that MobileBERT is 4.3$\times$ smaller and 5.5$\times$ faster than  $\text{BERT}_\text{BASE}$ while achieving competitive results on well-known benchmarks.
	On the natural language inference tasks of GLUE, $\text{MobileBERT}$ achieves a GLUE score of $77.7$  ($0.6$ lower than $\text{BERT}_\text{BASE}$), 
	and 62 ms latency on a Pixel 4 phone. On the SQuAD v1.1/v2.0 question answering task, $\text{MobileBERT}$ achieves a dev F1 score of $90.0/79.2$  ($1.5/2.1$ higher than $\text{BERT}_\text{BASE}$). 
	
\end{abstract}

\section{Introduction}

The NLP community has witnessed a revolution of pre-training self-supervised models. These models usually have hundreds of millions of parameters \citep{peters2018deep,radford2018improving,devlin2018bert, radford2019language,yang2019xlnet}.
Among these models, BERT \citep{devlin2018bert} shows substantial accuracy improvements.
However, as one of the largest models ever in NLP, BERT suffers from the heavy model size and high latency, making it impractical for resource-limited mobile devices to deploy the power of BERT in mobile-based machine translation, dialogue modeling, and the like.

There have been some efforts that task-specifically distill BERT into compact models  \citep{turc2019well,tang2019distilling,sun2019patient,tsai2019small}. To the best of our knowledge,  there is not yet any work for building a task-agnostic lightweight pre-trained model, that is, a model that can be generically fine-tuned on different downstream NLP tasks as the original BERT does. In this paper, we propose MobileBERT to fill this gap.  In practice,  task-agnostic compression of BERT is desirable.  Task-specific compression needs to first fine-tune the original large BERT model into a task-specific teacher and then distill. Such a process is much more complicated \citep{wu2019conditional} and costly than directly fine-tuning a task-agnostic compact model.

\begin{figure*}[t]
	\centering
	\includegraphics[width=0.8\linewidth]{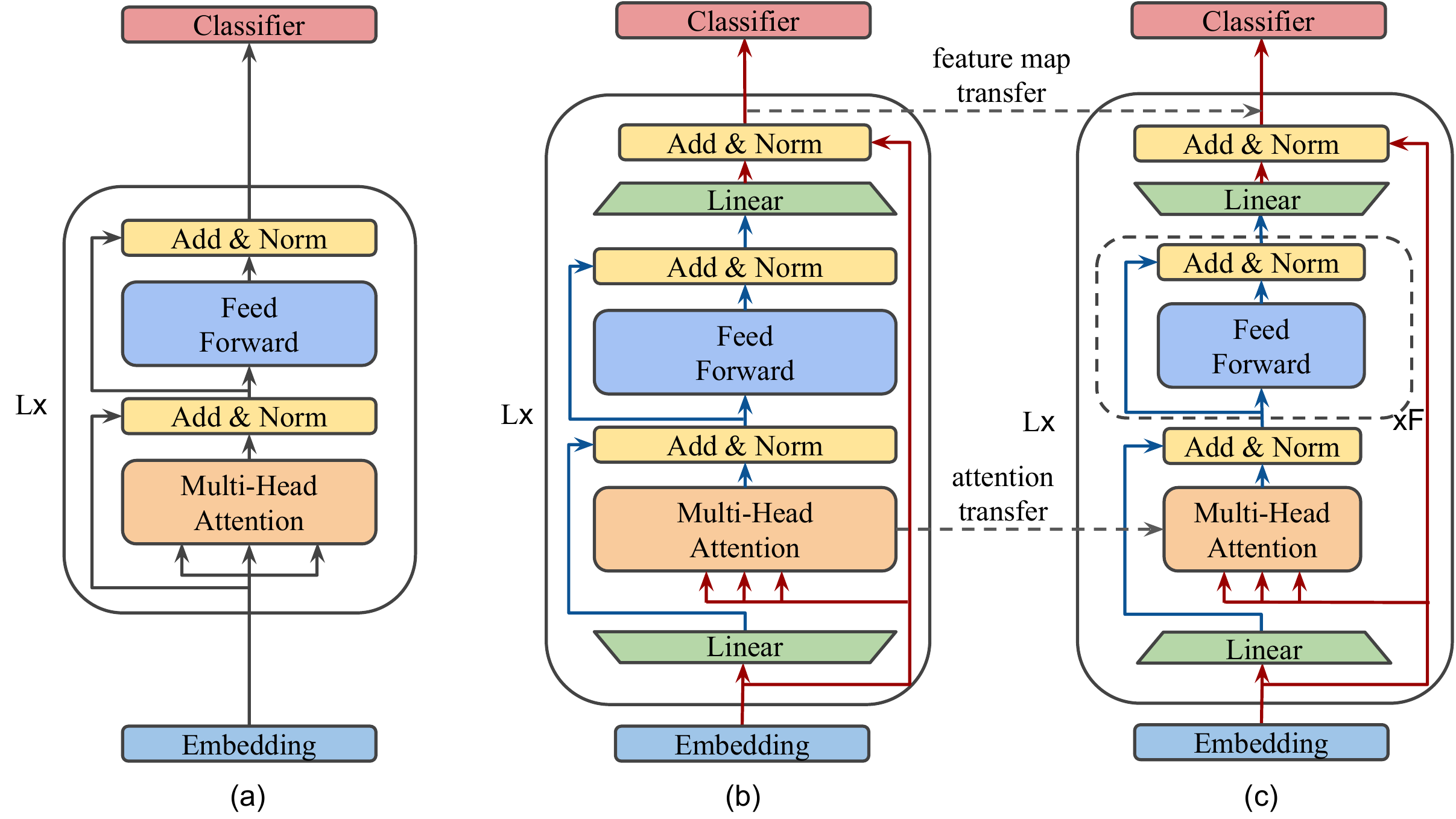}
\caption{Illustration of three models:  (a) BERT; (b) Inverted-Bottleneck BERT (IB-BERT); and (c) MobileBERT.  In (b) and (c),
	\textcolor{myred}{red lines denote {inter-block} flows} while \textcolor{myblue}{{blue lines} {intra-block} flows}. MobileBERT is  trained by layer-to-layer imitating IB-BERT.  
}
\label{fig:bottleneck}
\end{figure*}

At first glance, it may seem straightforward to obtain a task-agnostic compact BERT.  For example, one may just take a narrower or shallower version of BERT, and train it until convergence by minimizing a convex combination of the prediction loss and distillation loss \citep{turc2019well, sun2019patient}. Unfortunately, empirical results show that such a straightforward approach results in significant accuracy loss \citep{turc2019well}.  This may not be that surprising. It is well-known that shallow networks usually do not have enough representation power while narrow and deep networks are difficult to train. 

Our MobileBERT is designed to be as deep as $\text{BERT}_\text{LARGE}$ 
while each layer is made much narrower via adopting bottleneck structures and balancing between self-attentions and feed-forward networks (Figure \ref{fig:bottleneck}). To train MobileBERT, a deep and thin model,  we  
first train a specially designed teacher model, an inverted-bottleneck incorporated $\text{BERT}_\text{LARGE}$ model ($\text{IB-BERT}$). Then, we conduct knowledge transfer from IB-BERT to MobileBERT. A variety of knowledge transfer strategies are carefully investigated in our empirical studies.

Empirical evaluations\footnote{The code and pre-trained models are available at \url{https://github.com/google-research/google-research/tree/master/mobilebert}.} show that MobileBERT is 4.3$\times$ smaller and 5.5$\times$ faster than  $\text{BERT}_\text{BASE}, $ while it can still achieve competitive results on well-known NLP benchmarks.
On the natural language inference tasks of GLUE, MobileBERT can achieve a GLUE score of $77.7$, which is only $0.6$ lower than $\text{BERT}_\text{BASE}$, with a  latency of 62 ms on a Pixel 4 phone. On the SQuAD v1.1/v2.0 question answering task, MobileBER  obtains a dev F1 score  of $90.3/80.2$,   which is even $1.5/2.1$ higher than $\text{BERT}_\text{BASE}$.

\section{Related Work}

Recently, compression of BERT has attracted much attention. \citet{turc2019well} propose to pre-train the smaller BERT models to improve task-specific knowledge distillation. \citet{tang2019distilling} distill BERT into an extremely small LSTM model. \citet{tsai2019small} distill a multilingual BERT into smaller BERT models on sequence labeling tasks.
\citet{clark2019bam} use several single-task BERT models to teach a multi-task BERT. \citet{liu2019multi} distill knowledge from an ensemble of BERT models into a single BERT.

Concurrently to our work, \citet{sun2019patient} distill BERT into shallower students through knowledge distillation and an additional knowledge transfer of hidden states on multiple intermediate layers. \citet{jiao2019tinybert} propose TinyBERT, which also uses a layer-wise distillation strategy for BERT but in both pre-training and fine-tuning stages. \citet{sanh2019distilbert} propose DistilBERT, which successfully halves the depth of BERT model by knowledge distillation in the pre-training stage and an optional fine-tuning stage.

In contrast to these existing literature, we only use knowledge transfer in the pre-training stage and do not require a fine-tuned teacher or data augmentation \cite{wu2019conditional} in the down-stream tasks. Another key difference is that these previous work try to compress BERT by reducing its depth, while we focus on compressing BERT by reducing its width, which has been shown to be more effective \cite{turc2019well}.

\newcommand{\blockl}[4]{\multirow{10}{*}{
\(\left[
\begin{array}{c}
$\(\left(
\begin{array}{c}
\text{#1}\\
[-.1em] \text{16}\\
[-.1em] \text{#2}\\
\end{array}\right)\)$\\
[-.1em] $\(\left(
\begin{array}{c}
\text{#1}\\
[-.1em] \text{#4}\\
[-.1em] \text{#1}\\
\end{array}\right)\)$\\
\end{array}\right]\)$\times$#3}
}

\newcommand{\blockb}[4]{\multirow{10}{*}{
\(\left[
\begin{array}{c}
$\(\left(
\begin{array}{c}
\text{#1}\\
[-.1em] \text{12}\\
[-.1em] \text{#2}\\
\end{array}\right)\)$\\
[-.1em] $\(\left(
\begin{array}{c}
\text{#1}\\
[-.1em] \text{#4}\\
[-.1em] \text{#1}\\
\end{array}\right)\)$\\
\end{array}\right]\)$\times$#3}
}

\newcommand{\blockx}[4]{\multirow{10}{*}{
\(\left[
\begin{array}{l}
$\(\left(
\begin{array}{c}
\text{#1}\\
[-.1em] \text{#2}\\
\end{array}\right)\)$\\
[-.1em] $\(\left(
\begin{array}{c}
\text{#1}\\
[-.1em] \text{4}\\
[-.1em] \text{#2}\\
\end{array}\right)\)$\\
[-.1em] $\(\left(
\begin{array}{c}
\text{#2}\\
[-.1em] \text{#4}\\
[-.1em] \text{#2}\\
\end{array}\right)\)$\\
[-.1em] $\(\left(
\begin{array}{c}
\text{#2}\\
[-.1em] \text{#1}\\
\end{array}\right)\)$\\
\end{array}\right]\)$\times$#3}
}

\newcommand{\blockxx}[4]{\multirow{10}{*}{
\(\left[
\begin{array}{l}
$\(\left(
\begin{array}{c}
\text{#1}\\
[-.1em] \text{#2}\\
\end{array}\right)\)$\\
[-.1em] $\(\left(
\begin{array}{c}
\text{#1}\\
[-.1em] \text{4}\\
[-.1em] \text{#2}\\
\end{array}\right)\)$\\
[-.1em] $\(\left(
\begin{array}{c}
\text{#2}\\
[-.1em] \text{#4}\\
[-.1em] \text{#2}\\
\end{array}\right)\)$\times \text{4}\\
[-.1em] $\(\left(
\begin{array}{c}
\text{#2}\\
[-.1em] \text{#1}\\
\end{array}\right)\)$\\
\end{array}\right]\)$\times$#3}
}

\newcommand{\blockxxx}[4]{\multirow{10}{*}{
\(\left[
\begin{array}{l}
$\(\left(
\begin{array}{c}
\text{#1}\\
[-.1em] \text{#2}\\
\end{array}\right)\)$\\
[-.1em] $\(\left(
\begin{array}{c}
\text{#2}\\
[-.1em] \text{4}\\
[-.1em] \text{#2}\\
\end{array}\right)\)$\\
[-.1em] $\(\left(
\begin{array}{c}
\text{#2}\\
[-.1em] \text{#4}\\
[-.1em] \text{#2}\\
\end{array}\right)\)$\times \text{2}\\
[-.1em] $\(\left(
\begin{array}{c}
\text{#2}\\
[-.1em] \text{#1}\\
\end{array}\right)\)$\\
\end{array}\right]\)$\times$#3}
}

\newcolumntype{x}[1]{>\centering p{#1pt}}
\newcommand{\ft}[1]{\fontsize{#1pt}{1em}\selectfont}
\renewcommand\arraystretch{1.25}
\setlength{\tabcolsep}{1.2pt}
\begin{table*}[t]
\begin{center}
\resizebox{\linewidth}{!}{
\footnotesize
\begin{tabular}{c|x{40}|c|c|c|c|c|c}
\hline
  \multicolumn{3}{c|}{}& $\text{BERT}_\text{LARGE}$ & $\text{BERT}_\text{BASE}$ & $\text{IB-BERT}_\text{LARGE}$ & $\text{MobileBERT}$ & $\text{MobileBERT}_\text{TINY}$\\
\hline
\multicolumn{2}{c|}{\multirow{3}{*}{embedding}} & $\text{h}_\text{embedding}$ & 1024 & 768 & \multicolumn{3}{c}{128} \\
\cline{4-8}
\multicolumn{2}{c|}{} & & no-op & no-op & \multicolumn{3}{c}{3-convolution}\\
\cline{4-8}
\multicolumn{2}{c|}{} & $\text{h}_\text{inter}$ & 1024 & 768 & \multicolumn{3}{c}{512}\\
\hline

\multirow{10}{*}{body} & \multirow{2}{*}{Linear} & $\text{h}_\text{input}$ & \blockl{1024}{1024}{24}{4096}  & \blockb{768}{768}{12}{3072} & \blockx{512}{1024}{24}{4096} &
\blockxx{512}{128}{24}{512} &
\blockxxx{512}{128}{24}{512}\\
&  & $\text{h}_\text{output}$ &  & & &&\\
\cline{2-3}
&\multirow{3}{*}{MHA}  & $\text{h}_\text{input}$ &  & & &&\\
&  &  \ft{8} \#Head &  & & &&\\
&  & $\text{h}_\text{output}$  &  & & &&\\
\cline{2-3}
&\multirow{3}{*}{FFN}  & $\text{h}_\text{input}$  &  & && &\\
&  & $\text{h}_\text{FFN}$  &  & & &&\\
&  & $\text{h}_\text{output}$  &  & & &&\\
\cline{2-3}
&\multirow{2}{*}{Linear}  & $\text{h}_\text{input}$ &  & & &&\\
&  & $\text{h}_\text{output}$ &  & & & &\\
\hline

\multicolumn{3}{c|}{\small \#Params} & 334M & 109M & 293M & 25.3M & 15.1M  \\
\hline
\end{tabular}
}
\end{center}
\caption{
The detailed model settings of a few models. $\text{h}_\text{inter}$, $\text{h}_\text{FFN}$, $\text{h}_\text{embedding}$, \text{\#Head} and \text{\#Params}  denote the inter-block hidden size (feature map size),  FFN intermediate size, embedding table size, the number of heads in multi-head attention, and the number of parameters, respectively.
}
\label{tab:detail}
\vspace{-.5em}
\end{table*}

\section{MobileBERT}\label{sec:mobilebert}

In this section, we present the detailed architecture design of MobileBERT and training strategies to efficiently train MobileBERT. The specific model settings are summarized in Table \ref{tab:detail}. These settings are obtained by extensive architecture search experiments which will be presented in Section \ref{sec:nas}.

\subsection{Bottleneck and Inverted-Bottleneck}

The architecture of MobileBERT is illustrated in Figure 1(c). It is as deep as  $\text{BERT}_\text{LARGE}, $  but each building block is made much smaller. As shown in Table \ref{tab:detail}, the hidden dimension of each building block is only 128.  On the other hand, we introduce two linear transformations for each building block to adjust its input and output dimensions to 512. Following the terminology in \citep{he2016deep}, we refer to such an architecture  as  bottleneck. 

It is challenging to train such a deep and thin network. To overcome the training issue, we first construct a teacher network and train it until convergence, and then conduct knowledge transfer from this teacher network to MobileBERT.  We find that this is much better than directly training MobileBERT from scratch. Various training strategies will be discussed in a later section. Here, we introduce the architecture design of the teacher network which is illustrated in Figure 1(b). In fact, the teacher network is just  $\text{BERT}_\text{LARGE}$ while augmented with \emph{inverted}-bottleneck structures \citep{sandler2018mobilenetv2} to adjust its feature map size to 512. 
In what follows, we refer to the teacher network as $\text{IB-BERT}_\text{LARGE}$.
Note that IB-BERT and MobileBERT have the same feature map size which is 512. Thus, we can directly compare the layer-wise output difference between IB-BERT and MobileBERT. Such a direct comparison is needed in our knowledge transfer strategy.

It is worth pointing out that the simultaneously introduced bottleneck and inverted-bottleneck structures result in a fairly flexible architecture design. One may either only use the bottlenecks for MobileBERT (correspondingly the teacher becomes $\text{BERT}_\text{LARGE}$) or only the inverted-bottlenecks for IB-BERT (then there is no bottleneck in MobileBERT) to align their feature maps. However, when using both of them,  we can allow $\text{IB-BERT}_\text{LARGE}$ to preserve the performance of $\text{BERT}_\text{LARGE}$ while having MobileBERT sufficiently compact.

\subsection{Stacked Feed-Forward Networks}

A problem introduced by the bottleneck structure of MobileBERT is that the balance between the Multi-Head Attention (MHA) module and the Feed-Forward Network (FFN) module is broken.
MHA and FFN play different roles in the Transformer architecture: The former allows the model to jointly attend to information from different subspaces, while the latter increases the non-linearity of the model.
In original BERT, the ratio of the parameter numbers in MHA and FFN is always 1:2. But in the bottleneck structure, the inputs to the MHA are from wider feature maps (of inter-block size), while the inputs to the FFN are from narrower bottlenecks (of intra-block size). This results in that the MHA modules in MobileBERT relatively contain more parameters.

To fix this issue, we propose to use stacked feed-forward networks in MobileBERT to re-balance the relative size between MHA and FFN. As illustrated in Figure \ref{fig:bottleneck}(c), each MobileBERT layer contains one MHA but several stacked FFN. In MobileBERT, we use 4 stacked FFN after each MHA.

\subsection{Operational Optimizations}\label{sec:opt}
By model latency analysis\footnote{A detailed analysis of effectiveness of operational optimizations on real-world inference latency can be found in Section \ref{app:latency}.}, we find that layer normalization \citep{ba2016layer} and $\mathtt{gelu}$ activation \citep{hendrycks2016bridging} accounted for a considerable proportion of total latency. Therefore, we propose to replace them with new operations in our MobileBERT.
\paragraph{Remove layer normalization} We replace the layer normalization of a $n$-channel hidden state $\mathbf{h}$ with an element-wise linear transformation:
\begin{equation}
    \mathtt{NoNorm}(\mathbf{h}) = \boldsymbol{\gamma} \circ \mathbf{h} + \boldsymbol{\beta},
\end{equation}
where $\boldsymbol{\gamma}, \boldsymbol{\beta} \in \mathbb{R}^{n}$ 
and $\circ$ denotes the Hadamard product. Please note that $\mathtt{NoNorm}$ has different properties from $\mathtt{LayerNorm}$ even in test mode since the original layer normalization is not a linear operation for a batch of vectors.
\paragraph{Use relu activation} We replace the $\mathtt{gelu}$ activation with simpler $\mathtt{relu}$ activation \citep{nair2010rectified}.

\makeatletter
\newcommand{\removelatexerror}{\let\@latex@error\@gobble}
\makeatother

\newcommand\mycommfont[1]{\scriptsize\ttfamily{#1}}
\SetCommentSty{mycommfont}

\begin{figure*}[t]
\begin{small}
\centering
\includegraphics[width=\linewidth]{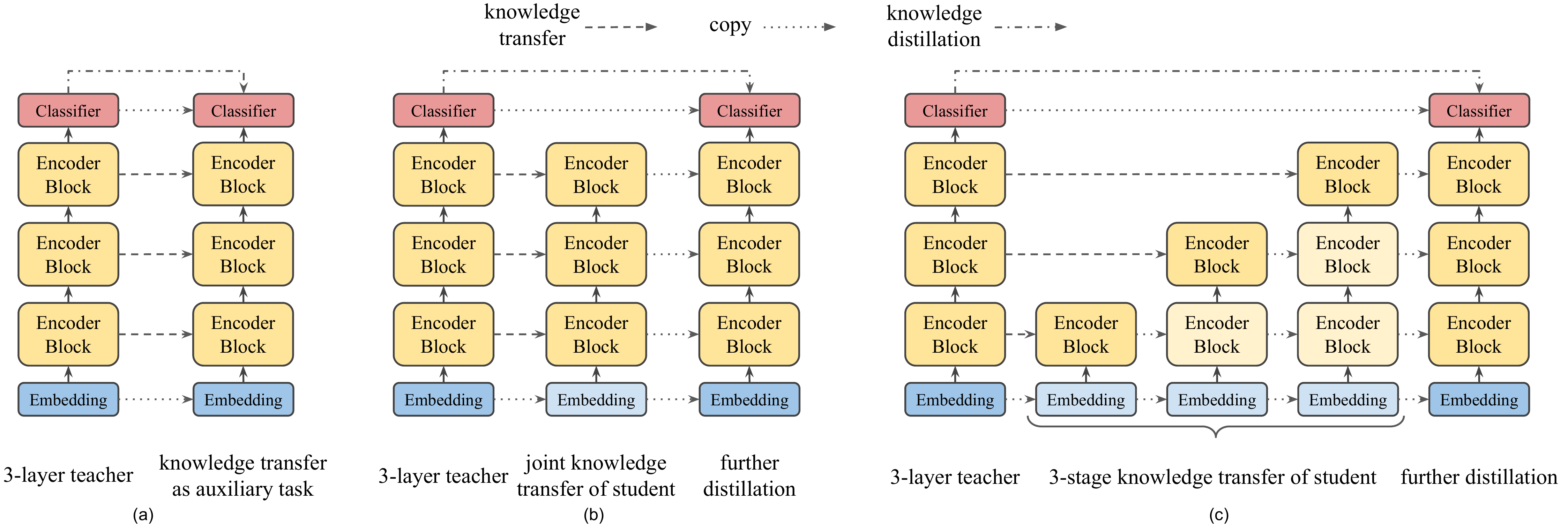}
\end{small}
  \vspace{-1.5em}
  \caption{Diagrams of (a) auxiliary knowledge transfer (AKT), (b) joint knowledge transfer (JKT), and (c) progressive knowledge transfer (PKT). Lighter colored blocks represent that they are frozen in that stage. 
}
\label{fig:diagram}
\end{figure*}

\subsection{Embedding Factorization}

The embedding table in BERT models accounts for a substantial proportion of model size. To compress the embedding layer, as shown in Table \ref{tab:detail}, 
we reduce the embedding dimension to 128 in MobileBERT. Then, we apply a 1D convolution with kernel size 3 on the raw token embedding to produce a 512 dimensional output.

\subsection{Training Objectives}

We propose to use the following two knowledge transfer objectives, i.e., feature map transfer and attention transfer, to train MobileBERT. 
Figure \ref{fig:bottleneck} illustrates the proposed layer-wise knowledge transfer objectives. Our final layer-wise knowledge transfer loss $\mathcal{L}_{KT}^\ell$ for the $\ell^{th}$ layer is a linear combination of the two objectives stated below:

\paragraph{Feature Map Transfer (FMT)}
Since each layer in BERT merely takes the output of the previous layer as input, the most important thing in layer-wise knowledge transfer is that the feature maps of each layer should be as close as possible to those of the teacher. In particular, the mean squared error between the feature maps of the MobileBERT student and the IB-BERT teacher is used as the knowledge transfer objective:
\useshortskip
\begin{equation}
\mathcal{L}_{FMT}^\ell = \frac{1}{TN}\sum_{t=1}^T\sum_{n=1}^N(H^{tr}_{t,\ell,n}-H^{st}_{t,\ell,n})^2,
\end{equation}
where $\ell$ is the index of layers, $T$ is the sequence length, and $N$ is the feature map size.
In practice, we find that decomposing this loss term into normalized feature map discrepancy and feature map statistics discrepancy can help stabilize training.

\paragraph{Attention Transfer (AT)}

The attention mechanism greatly boosts the performance of NLP and becomes a crucial building block in Transformer and BERT \citep{clark2019does,jawahar2019does}.
This motivates us to use self-attention maps from the well-optimized teacher to help the training of MobileBERT in augmentation to the feature map transfer.
In particular, we minimize the KL-divergence between the per-head self-attention distributions of the MobileBERT student and the IB-BERT teacher:
\begin{equation}
\mathcal{L}_{AT}^\ell = \frac{1}{TA}\sum_{t=1}^T\sum_{a=1}^A D_{KL}(a_{t,\ell,a}^{tr}||a_{t,\ell,a}^{st}),
\end{equation}
where $A$ is the number of attention heads.

\paragraph{Pre-training Distillation (PD)}

Besides layer-wise knowledge transfer, we can also use a knowledge distillation loss when pre-training MobileBERT.
We use a linear combination of the original masked language modeling (MLM) loss, next sentence prediction (NSP) loss, 
and the new MLM Knowledge Distillation (KD) loss as our pre-training distillation loss:
\begin{equation}
\mathcal{L}_{PD} = \alpha \mathcal{L}_{MLM} + (1 - \alpha) \mathcal{L}_{KD} + \mathcal{L}_{NSP},
\end{equation}
where $\alpha$ is a hyperparameter in $(0, 1)$.

\subsection{Training Strategies} \label{sec:bottom}

Given the objectives defined above, there can be various combination strategies in training. We discuss three strategies in this paper.

\paragraph{Auxiliary Knowledge Transfer}

In this strategy, we regard intermediate knowledge transfer as an auxiliary task for knowledge distillation. We use a single loss, which is a linear combination of knowledge transfer losses from all layers as well as the pre-training distillation loss.

\paragraph{Joint Knowledge Transfer}

However, the intermediate knowledge of the IB-BERT teacher (i.e. attention maps and feature maps) may not be an optimal solution for the MobileBERT student.
Therefore, we propose to separate these two loss terms, where we first train MobileBERT with all layer-wise knowledge transfer losses jointly, and then further train it by pre-training distillation.

\paragraph{Progressive Knowledge Transfer}

One may also concern that if MobileBERT cannot perfectly mimic the IB-BERT teacher, the errors from the lower layers may affect the knowledge transfer in the higher layers. Therefore, we propose to progressively train each layer in the knowledge transfer. The progressive knowledge transfer is divided into $L$ stages, where $L$ is the number of layers.

\paragraph{Diagram of three strategies}

Figure \ref{fig:diagram} illustrates the diagram of the three strategies. For joint knowledge transfer and progressive knowledge transfer, there is no knowledge transfer for the beginning embedding layer and the final classifier in the layer-wise knowledge transfer stage. They are copied from the IB-BERT teacher to the MobileBERT student. Moreover, for progressive knowledge transfer, when we train the $\ell^{th}$ layer,  we freeze all the trainable parameters in the layers below. In practice, we can soften the training process as follows. When training a layer, we further tune the lower layers with a small learning rate rather than entirely freezing them.

\section{Experiments}
 
In this section, we first present our architecture search experiments which lead to the model settings in Table \ref{tab:detail}, and then present the empirical results on benchmarks from MobileBERT and various baselines. 

\subsection{Model Settings} \label{sec:nas} 

We conduct extensive experiments to search good model settings for the IB-BERT teacher and the MobileBERT student.
We start with SQuAD v1.1 dev F1 score as the performance metric in the search of model settings. In this section, we only train each model for 125k steps with 2048 batch size, which halves the training schedule of original BERT \citep{devlin2018bert,you2019large}.

\setlength{\tabcolsep}{2pt}

\begin{table}[t]
\centering
	    \small
		\begin{tabular}{l c c c c c}
			\hline
			& \textbf{\#Params} & $\mathbf{h_{inter}}$ & $\mathbf{h_{intra}}$ & \textbf{\#Head} & \textbf{SQuAD}\\
			\hline
			(a) & 356M & 1024 & 1024 & 16 & 88.2 \\
			(b) & 325M & 768 & 1024 & 16 & 88.6 \\
			(c) & 293M & 512 & 1024 & 16 & 88.1 \\
			(d) & 276M & 384 & 1024 & 16 & 87.6 \\
			(e) & 262M & 256 & 1024 & 16 & 87.0 \\
			\hline
			(f) & 293M & 512 & 1024 & 4 & 88.3 \\
			(g) & 92M & 512 & 512 & 4 & 85.8 \\
			(h) & 33M & 512 & 256 & 4 & 84.8 \\
            (i) & 15M & 512 & 128 & 4 & 82.0 \\
			\hline
		\end{tabular}
\caption{Experimental results on SQuAD v1.1 dev F1 score in search of good model settings for the $\text{IB-BERT}_\text {LARGE}$ teacher.  The number of layers is set to 24 for all models.}
    \label{tab:bottleneck}
\end{table}
\setlength{\tabcolsep}{6pt}

\paragraph{Architecture Search for IB-BERT}

Our design philosophy for the teacher model is to use as small inter-block hidden size (feature map size) as possible, as long as there is no accuracy loss. Under this guideline, we design experiments to manipulate the inter-block size of a $\text{BERT}_\text{LARGE}$-sized IB-BERT, and the results are shown in Table \ref{tab:bottleneck} with labels (a)-(e).
We can see that reducing the inter-block hidden size doesn't damage the performance of BERT until it is smaller than 512. Hence, we choose $\text{IB-BERT}_\text{LARGE}$ with its inter-block hidden size being 512 as the teacher model.

One may wonder whether we can also shrink the intra-block hidden size of the teacher. We conduct experiments and the results are shown in Table \ref{tab:bottleneck} with labels (f)-(i).  We can see that when the intra-block hidden size is reduced, the model performance is dramatically worse. This means that the intra-block hidden size, which represents the representation power of non-linear modules, plays a crucial role in BERT. 
Therefore, unlike the inter-block hidden size, we do not shrink the intra-block hidden size of our teacher model.

\setlength{\tabcolsep}{2pt}
\begin{table}[t]
    \centering
    \small
	\begin{tabular}{c c c c c}
		\hline
		$\mathbf{h_{intra}}$  & \textbf{\#Head (\#Params)} & \textbf{\#FFN (\#Params)} & \textbf{SQuAD}\\
		\hline
		192 & 6 \ \ \ (8M) & 1 \ \ \ \ \ (7M)  & 82.6 \\
		160 & 5 (6.5M) & 2 \ \ \ (10M)  & 83.4 \\			
		128 & 4 \ \ \ (5M) & 4 (12.5M)  & 83.4 \\
		96 & 3 \ \ \ (4M) & 8 \ \ \ (14M)  & 81.6 \\
		\hline
\end{tabular}
	\caption{Experimental results on SQuAD v1.1 dev F1 score in search of good model settings for the MobileBERT student. The number of layers is set to 24 and the inter-block hidden size is set to 512 for all models. }
	\label{tab:bottleneck2}
\end{table}
\setlength{\tabcolsep}{6pt}

\setlength{\tabcolsep}{2pt}

\begin{table*}[t]
\begin{center}
	\resizebox{\linewidth}{!}{
	    \small
		\begin{tabular}{l | c c c| c c c c c c c c | c}
			\hline
            &\multirow{2}{*}{\textbf{\#Params}} & \multirow{2}{*}{\textbf{\#FLOPS}} & \multirow{2}{*}{\textbf{Latency}} & \textbf{CoLA} & \textbf{SST-2} & \textbf{MRPC} & \textbf{STS-B} & \textbf{QQP} & \textbf{MNLI-m/mm} & \textbf{QNLI} & \textbf{RTE} & \multirow{2}{*}{\textbf{GLUE}}\\
            & & & & 8.5k & 67k & 3.7k & 5.7k & 364k & 393k & 108k & 2.5k &\\
			\hline
            ELMo-BiLSTM-Attn & - & - & - & 33.6 & 90.4 & 84.4 & 72.3 & 63.1 & 74.1/74.5 & 79.8 & 58.9 & 70.0\\
            OpenAI GPT & 109M & - &- & 47.2 & 93.1 & 87.7 & 84.8 & 70.1 & 80.7/80.6 & 87.2 & 69.1 & 76.9\\
            $\text{BERT}_\text{BASE}$ & 109M & 22.5B & 342 ms & \textbf{52.1} & \textbf{93.5} & \textbf{88.9} & \textbf{85.8} & 71.2 & \textbf{84.6}/\textbf{83.4} & 90.5 & 66.4 & 78.3 \\
            $\text{BERT}_\text{BASE}$-6L-PKD* & 66.5M & 11.3B &- & - & 92.0 & 85.0 & - & 70.7 & 81.5/81.0 & 89.0 & 65.5 & -\\
            $\text{BERT}_\text{BASE}$-4L-PKD\dag* & 52.2M & 7.6B &- & 24.8 & 89.4 & 82.6 & 79.8 & 70.2 & 79.9/79.3 & 85.1 & 62.3 & - \\
            $\text{BERT}_\text{BASE}$-3L-PKD* & 45.3M & 5.7B &- & - & 87.5 & 80.7 & - & 68.1 & 76.7/76.3 & 84.7 & 58.2 & -\\
$\text{DistilBERT}_\text{BASE}$-6L\dag & 62.2M & 11.3B &- & - & 92.0 & 85.0 & & 70.7 & 81.5/81.0 & 89.0 & 65.5 & - \\
            $\text{DistilBERT}_\text{BASE}$-4L\dag & 52.2M & 7.6B &- & 32.8 & 91.4 & 82.4 & 76.1 & 68.5 & 78.9/78.0 & 85.2 & 54.1 & - \\
$\text{TinyBERT}$* & 14.5M & 1.2B &- & 43.3 & 92.6 & 86.4 & 79.9 & \textbf{71.3} & 82.5/81.8 & 87.7 & 62.9 & 75.4\\
			\hline
			$\text{MobileBERT}_\text{TINY}$ & 15.1M & 3.1B &40 ms & 46.7 & 91.7 & 87.9 & 80.1 & 68.9 & 81.5/81.6 & 89.5 & 65.1 & 75.8\\
			MobileBERT & 25.3M & 5.7B &62 ms & 50.5 & 92.8 & 88.8 & 84.4 & 70.2 & 83.3/82.6 & 90.6  & 66.2  &  77.7 \\
			MobileBERT w/o OPT& 25.3M & 5.7B &192 ms & 51.1 & 92.6 & 88.8 & 84.8 & 70.5 & 84.3/\textbf{83.4} & \textbf{91.6} & \textbf{70.4} & \textbf{78.5} \\
			\hline
		\end{tabular}
	}
	\end{center}
	\caption{The test results on the GLUE benchmark (except WNLI). The number below each task denotes the number of training examples. The metrics for these tasks can be found in the GLUE paper \citep{wang2018glue}. ``OPT'' denotes the operational optimizations introduced in Section \ref{sec:opt}. \dag denotes that the results are taken from \cite{jiao2019tinybert}. *denotes that it can be unfair to directly compare MobileBERT with these models since MobileBERT is task-agnosticly compressed while these models use the teacher model in the fine-tuning stage.}
	\label{tab:glue}
\end{table*}
\setlength{\tabcolsep}{6pt}

\paragraph{Architecture Search for MobileBERT}

We seek a compression ratio of 4$\times$ for $\text{BERT}_\text{BASE}$, so we design a set of MobileBERT models all with approximately 25M parameters but different ratios of the parameter numbers in MHA and FFN to select a good MobileBERT student model. Table \ref{tab:bottleneck2} shows our experimental results.
They have different balances between MHA and FFN. 
From the table, we can see that the model performance reaches the peak when the ratio of parameters in MHA and FFN is 0.4 $\sim$ 0.6. This may justify why the original Transformer chooses the parameter ratio of MHA and FFN to 0.5. 

We choose the architecture with 128 intra-block hidden size and 4 stacked FFNs as the MobileBERT student model in consideration of model accuracy and training efficiency. We also accordingly set the number of attention heads in the teacher model to 4 in preparation for the layer-wise knowledge transfer. Table \ref{tab:detail} demonstrates the model settings of our $\text{IB-BERT}_\text{LARGE}$ teacher and MobileBERT student.

One may wonder whether reducing the number of heads will harm the performance of the teacher model. By comparing (a) and (f) in Table \ref{tab:bottleneck}, we can see that reducing the number of heads from 16 to 4 does not affect the performance of $\text{IB-BERT}_\text{LARGE}$.

\subsection{Implementation Details}

Following BERT \citep{devlin2018bert}, we use the BooksCorpus \citep{zhu2015aligning} and English Wikipedia as our pre-training data. To make the $\text{IB-BERT}_\text{LARGE}$ teacher reach the same accuracy as original $\text{BERT}_\text{LARGE}$, we train $\text{IB-BERT}_\text{LARGE}$ on 256 TPU v3 chips for 500k steps with a batch size of 4096 and LAMB optimizer \citep{you2019large}. For a fair comparison with the original BERT, we do not use training tricks in other BERT variants \citep{liu2019roberta,joshi2019spanbert}.
For MobileBERT, we use the same training schedule in the pre-training distillation stage. Additionally, we use progressive knowledge transfer to train MobileBERT, which takes additional 240k steps over 24 layers.
In ablation studies, we halve the pre-training distillation schedule of MobileBERT to accelerate experiments. Moreover, in the ablation study of knowledge transfer strategies, for a fair comparison, joint knowledge transfer and auxiliary knowledge transfer also take additional 240k steps.

For the downstream tasks, all reported results are obtained by \textbf{simply fine-tuning 
MobileBERT just like what the original BERT does}. To fine-tune the pre-trained models, we search the optimization hyperparameters in a search space including different batch sizes (16/32/48), learning rates ((1-10) * e-5), and the number of epochs (2-10). The search space is different from the original BERT because we find that MobileBERT usually needs a larger learning rate and more training epochs in fine-tuning. We select the model for testing according to their performance on the development (dev) set.

\subsection{Results on GLUE}

The General Language Understanding Evaluation (GLUE) benchmark \citep{wang2018glue} is a collection of 9 natural language understanding tasks.
We compare MobileBERT with $\text{BERT}_\text{BASE}$ and a few state-of-the-art pre-BERT models on the GLUE leaderboard\footnote{\url{https://gluebenchmark.com/leaderboard}}: OpenAI GPT \citep{radford2018improving} and ELMo \citep{peters2018deep}. We also compare with three recently proposed compressed BERT models: BERT-PKD \citep{sun2019patient}, and DistilBERT \cite{sanh2019distilbert}.
To further show the advantage of MobileBERT over recent small BERT models, we also evaluate a smaller variant of our model with approximately 15M parameters called $\text{MobileBERT}_\text{TINY}$\footnote{The detailed model setting of $\text{MobileBERT}_\text{TINY}$ can be found in Table \ref{tab:detail} and in the appendix.}, which reduces the number of FFNs in each layer and uses a lighter MHA structure.
Besides, to verify the performance of MobileBERT on real-world mobile devices, we export the models with TensorFlow Lite\footnote{\url{https://www.tensorflow.org/lite}} APIs and measure the inference latencies on a 4-thread Pixel 4 phone with a fixed sequence length of 128. The results are listed in Table \ref{tab:glue}.
\footnote{We follow \citet{devlin2018bert} to skip the WNLI task. }

\setlength{\tabcolsep}{2pt}

\begin{table}[t]
\begin{center}
	\resizebox{\columnwidth}{!}{
	    \small
		\begin{tabular}{ l | c | c c c c }
			\hline
& \multirow{2}{*}{\textbf{\#Params}} & \multicolumn{2}{c}{\textbf{SQuAD v1.1}} &  \multicolumn{2}{c}{\textbf{SQuAD v2.0}}\\
			 &&  \ \ \textbf{EM} \ \  & \textbf{F1} & \ \  \textbf{EM} \ \  & \textbf{F1}\\
			\hline
			DocQA + ELMo &
			- &
			- & - & 65.1 & 67.6\\
            $\text{BERT}_\text{BASE}$ & 
109M &
            80.8 & 88.5 & 74.2\dag & 77.1\dag \\
            $\text{DistilBERT}_\text{BASE}$-6L & 
66.6M &
            79.1 & 86.9 & - & - \\
            $\text{DistilBERT}_\text{BASE}$-6L\ddag & 
66.6M &
            78.1 & 86.2 & 66.0 & 69.5 \\
            $\text{DistilBERT}_\text{BASE}$-4L\ddag &
52.2M &
            71.8 & 81.2 & 60.6 & 64.1 \\
$\text{TinyBERT}$ &
            14.5M &
            72.7 & 82.1 & 65.3 & 68.8 \\
\hline
            $\text{MobileBERT}_\text{TINY}$ &
            15.1M &
            81.4 & 88.6 & 74.4 & 77.1 \\
            MobileBERT &
            25.3M &
82.9 & 90.0 & 76.2 & 79.2 \\
            MobileBERT w/o OPT& 
            25.3M &
\textbf{83.4} & \textbf{90.3} & \textbf{77.6} & \textbf{80.2} \\
            \hline
		\end{tabular}
    }
	\end{center}
	\caption{The results on the SQuAD dev datasets. \dag marks our runs with the official code. \ddag denotes that the results are taken from \cite{jiao2019tinybert}.}
	\label{tab:squad}
\end{table}
\setlength{\tabcolsep}{6pt}

\setlength{\tabcolsep}{2pt}
\begin{table}[t]
\begin{center}
	\resizebox{\columnwidth}{!}{
	    \small
		\begin{tabular}{l | c c c c c}
			\hline
             & \textbf{MNLI-m} & \textbf{QNLI} & \textbf{MRPC} & \textbf{SST-2} & \textbf{SQuAD}\\
             \hline
			$\text{MobileBERT}_\text{TINY}$ &  \textbf{82.0} & \textbf{89.9} & \textbf{86.7} & \textbf{91.6} & \textbf{88.6}\\
			\quad + Quantization & \textbf{82.0} & 89.8 & 86.3 & \textbf{91.6} & 88.4\\
			\hline
			 MobileBERT & \textbf{83.9} & \textbf{91.0} &\textbf{87.5} & \textbf{92.1} & \textbf{90.0}\\
			 \quad + Quantization & \textbf{83.9} & 90.8 & 87.0 & 91.9 & \textbf{90.0}\\
			\hline
		\end{tabular}
    }
	\end{center}
	\caption{Results of MobileBERT on GLUE dev accuracy and SQuAD v1.1 dev F1 score with 8-bit Quantization.}
	\label{tab:quantization}
\end{table}
\setlength{\tabcolsep}{6pt}

From the table, we can see that MobileBERT is very competitive on the GLUE benchmark. MobileBERT achieves an overall GLUE score of 77.7, which is only 0.6 lower than $\text{BERT}_\text{BASE}$, while being 4.3$\times$ smaller and 5.5$\times$ faster than $\text{BERT}_\text{BASE}$. Moreover, It outperforms the strong OpenAI GPT baseline by 0.8 GLUE score with $4.3\times$ smaller model size. It also outperforms all the other compressed BERT models with smaller or similar model sizes.
Finally, we find that the introduced operational optimizations hurt the model performance a bit. Without these optimizations, MobileBERT can even outperforms $\text{BERT}_\text{BASE}$ by 0.2 GLUE score.

\subsection{Results on SQuAD}

SQuAD is a large-scale reading comprehension datasets. SQuAD1.1 \citep{rajpurkar2016squad} only contains questions that always have an answer in the given context, while SQuAD2.0 \citep{rajpurkar2018know} contains unanswerable questions. 
We evaluate MobileBERT only on the SQuAD dev datasets, as there is nearly no single model submission on SQuAD test leaderboard. We compare our MobileBERT with $\text{BERT}_\text{BASE}$, DistilBERT, and a strong baseline DocQA \citep{clark2017simple}. As shown in Table \ref{tab:squad}, MobileBERT outperforms a large margin over all the other models with smaller or similar model sizes.

\subsection{Quantization}

We apply the standard post-training quantization in TensorFlow Lite to MobileBERT. The results are shown in Table \ref{tab:quantization}. We find that while quantization can further compress MobileBERT by 4$\times$, there is nearly no performance degradation from it. This indicates that there is still a big room in the compression of MobileBERT.

\subsection{Ablation Studies}

\begin{table}[t]
	\begin{center}
	    \small
		\begin{tabular}{ c  c  c }
		    \hline
		    \textbf{Setting} & \textbf{\#FLOPS} & \textbf{Latency}\\
			\hline
             $\mathtt{LayerNorm}$ \& $\mathtt{gelu}$ & 5.7B & 192 ms\\
             $\mathtt{LayerNorm}$ \& $\mathtt{relu}$ & 5.7B & 167 ms\\
             $\mathtt{NoNorm}$ \& $\mathtt{gelu}$ & 5.7B & 92 ms\\
             $\mathtt{NoNorm}$ \& $\mathtt{relu}$ & 5.7B & 62 ms\\
            \hline
		\end{tabular}
	\end{center}
	\caption{The effectiveness of operational optimizations on real-world inference latency for MobileBERT.}
	\label{tab:latency}
\end{table}

\subsubsection{Operational Optimizations}\label{app:latency}

 We evaluate the effectiveness of the two operational optimizations introduced in Section \ref{sec:opt}, i.e., replacing layer normalization ($\mathtt{LayerNorm}$) with $\mathtt{NoNorm}$ and replacing $\mathtt{gelu}$ activation with $\mathtt{relu}$ activation. We report the inference latencies using the same experimental setting as in Section \ref{app:latency}. From Table \ref{tab:latency}, we can see that both $\mathtt{NoNorm}$ and $\mathtt{relu}$ are very effective in reducing the latency of MobileBERT, while the two operational optimizations do not reduce FLOPS. This reveals the gap between the real-world inference latency and the theoretical computation overhead (i.e., FLOPS).

\setlength{\tabcolsep}{3pt}
\begin{table}[t]

\begin{center}
	    \small
		\begin{tabular}{c | c c c c c}
			\hline
             & \textbf{MNLI-m} & \textbf{QNLI} & \textbf{MRPC} & \textbf{SST-2} & \textbf{SQuAD}\\
             \hline
			 AKT & 83.0 & 90.3 & 86.8 & 91.9 & 88.2\\
			 JKT & 83.5 & 90.5 & \textbf{87.5} & 92.0 & 89.7\\
			 PKT & \textbf{83.9} & \textbf{91.0} &\textbf{87.5} & \textbf{92.1} & \textbf{90.0}\\
			\hline
		\end{tabular}
	\end{center}
\caption{Ablation study of MobileBERT on GLUE dev accuracy and SQuAD v1.1 dev F1 score with Auxiliary Knowledge Transfer (AKT), Joint Knowledge Transfer (JKT), and Progressive Knowledge Transfer (PKT).}
\label{tab:strategy}
\end{table}

\setlength{\tabcolsep}{6pt}

\subsubsection{Training Strategies}

We also study how the choice of training strategy, i.e., auxiliary knowledge transfer, joint knowledge transfer, and progressive knowledge transfer, can affect the performance of MobileBERT. As shown in Table \ref{tab:strategy}, progressive knowledge transfer consistently outperforms the other two strategies. We notice that there is a significant performance gap between auxiliary knowledge transfer and the other two strategies. We think the reason is that the intermediate layer-wise knowledge (i.e., attention maps and feature maps) from the teacher may not be optimal for the student, so the student needs an additional pre-training distillation stage to fine-tune its parameters.

\setlength{\tabcolsep}{3pt}
\begin{table}[t]
\begin{center}
	    \small
		\begin{tabular}{l | c c c c}
			\hline
            & \textbf{MNLI-m} & \textbf{QNLI} & \textbf{MRPC} & \textbf{SST-2}\\
			\hline
			$\text{BERT}_\text{LARGE}$ & 86.6 & 92.1\dag & \textbf{87.8} & 93.7\\
			$\text{IB-BERT}_\text{LARGE}$ & \textbf{87.0} & \textbf{93.2} & 87.3 & \textbf{94.1}\\
			\hline
			$\text{BERT}_\text{BASE}$ & \textbf{84.4} & 91.1\dag & 86.7 & \textbf{92.9}\\
MobileBERT (bare)  & 80.8 & 88.2 & 84.3 & 90.1\\
			 \quad + PD & 81.1 & 88.9 & 85.5 & 91.7 \\
			 \quad + PD + FMT  & 83.8 & 91.1 & \textbf{87.0} & 92.2 \\
			 \quad + PD + FMT + AT & \textbf{84.4} & \textbf{91.5} & \textbf{87.0} & 92.5\\
\hline
		\end{tabular}
	\end{center}
	\caption[Caption for LOF]{Ablation on the dev sets of GLUE benchmark.  $\text{BERT}_\text{BASE}$ and the bare MobileBERT (i.e., w/o PD, FMT, AT, FMT \& OPT) use the standard BERT pre-training scheme. PD, AT, FMT, and OPT denote Pre-training Distillation, Attention Transfer, Feature Map Transfer, and operational OPTimizations respectively. \dag marks our runs with the official code. }
	\label{tab:ablation}
\end{table}
\setlength{\tabcolsep}{6pt}

\subsubsection{Training Objectives}\label{sec:ablation}

\setlist[itemize]{leftmargin=*,noitemsep, topsep=0pt}

We finally conduct a set of ablation experiments with regard to Attention Transfer (AT), Feature Map Transfer (FMT) and Pre-training Distillation (PD). The operational OPTimizations (OPT) are removed in these experiments to make a fair comparison between MobileBERT and the original BERT. The results are listed in Table \ref{tab:ablation}.

We can see that the proposed Feature Map Transfer contributes most to the performance improvement of MobileBERT, while Attention Transfer and Pre-training Distillation also play positive roles.
We can also find that our $\text{IB-BERT}_\text{LARGE}$ teacher is as powerful as the original $\text{IB-BERT}_\text{LARGE}$ while MobileBERT degrades greatly when compared to its teacher. So we believe that there is still a big room in the improvement of MobileBERT.

\section{Conclusion}

We have presented MobileBERT which is a task-agnostic compact variant of BERT. Empirical results on popular NLP benchmarks show that MobileBERT is comparable with $\text{BERT}_\text{BASE}$ while being much smaller and faster. MobileBERT can enable various NLP applications\footnote{\url{https://tensorflow.org/lite/models/bert_qa/overview}} to be easily deployed on mobile devices.

In this paper, we show that 1) it is crucial to keep MobileBERT deep and thin, 2) bottleneck/inverted-bottleneck structures enable effective layer-wise knowledge transfer, and 3) progressive knowledge transfer can efficiently train MobileBERT. We believe our findings are generic and can be applied to other model compression problems.

\bibliography{anthology,acl2020,iclr2020_conference}
\bibliographystyle{acl_natbib}

\appendix

\begin{center}
{\bf \large{
    Appendix for ``MobileBERT: a Compact Task-Agnostic BERT for Resource-Limited Devices''}}
\end{center}

\section{Extra Related Work on Knowledge Transfer}

Exploiting knowledge transfer to compress model size was first proposed by \citet{bucilua2006model}. 
The idea was then adopted in knowledge distillation \citep{hinton2015distilling}, which requires the smaller student network to mimic the class distribution output of the larger teacher network. Fitnets \citep{romero2014fitnets} make the student mimic the intermediate hidden layers of the teacher to train narrow and deep networks. \citet{luo2016face} show that the knowledge of the teacher can also be obtained from the neurons in the top hidden layer. Similar to our proposed progressive knowledge transfer scheme, \citet{yeo2018sequential} proposed a sequential knowledge transfer scheme to distill knowledge from a deep teacher into a shallow student in a sequential way. \citet{zagoruyko2016paying} proposed to transfer the attention maps of the teacher on images. \citet{li2018hint} proposed to transfer the similarity of hidden states and word alignment from an autoregressive Transformer teacher to a non-autoregressive student.

\begin{figure*}[t]
	\centering
	\makebox[\linewidth][c]{\includegraphics[width=0.8\linewidth]{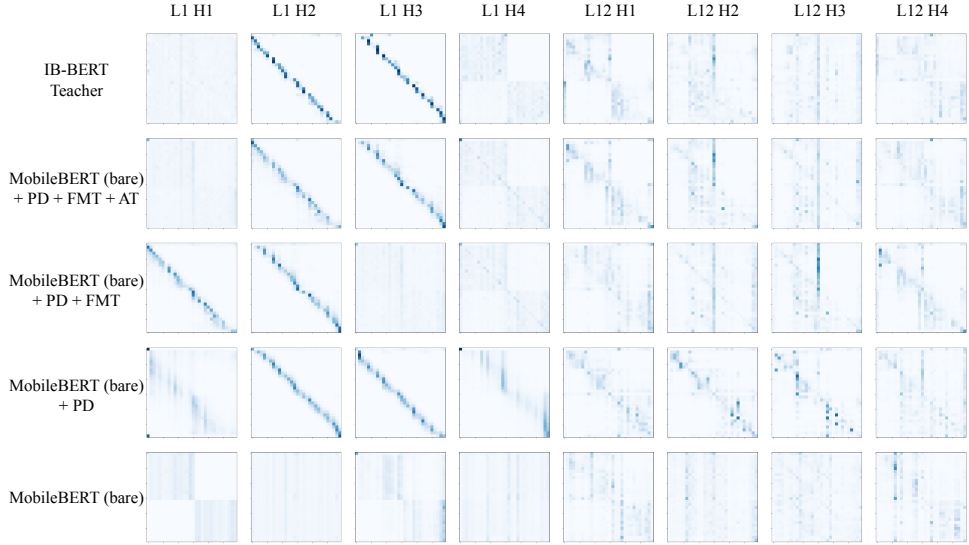}}
	\caption{The visualization of the attention distributions in some attention heads of the IB-BERT teacher and different MobileBERT models.}
	\label{fig:vis}
\end{figure*}

\section{Extra Related Work on Compact Architecture Design}
While much recent research has focused on improving efficient Convolutional Neural Networks (CNN) for mobile vision applications
\citep{iandola2016squeezenet,howard2017mobilenets,zhang2017interleaved,zhang2018shufflenet,sandler2018mobilenetv2,tan2019mnasnet,howard2019searching}, they are usually tailored for CNN.
Popular lightweight operations such as depth-wise convolution \citep{howard2017mobilenets} cannot be directly applied to Transformer or BERT. In the NLP literature, the most relevant work  can be group LSTMs \citep{kuchaiev2017factorization,gao2018efficient}, which employs the idea of group convolution \citep{zhang2017interleaved,zhang2018shufflenet} into Recurrent Neural Networks (RNN).

\section{Visualization of Attention Distributions}

We visualize the attention distributions of the $1^{st}$ and the $12^{th}$ layers of a few models in the ablation study for further investigation. They are shown in Figure \ref{fig:vis}. We find that the proposed attention transfer can help the student mimic the attention distributions of the teacher very well. Surprisingly, we find that the attention distributions in the attention heads of "MobileBERT(bare)+PD+FMT" are exactly a re-order of those of "MobileBERT(bare)+PD+FMT+AT" (also the teacher model), even if it has not been trained by the attention transfer objective. This phenomenon indicates that multi-head attention is a crucial and unique part of the non-linearity of BERT. Moreover, it can explain the minor improvements of Attention Transfer in the ablation study table, since \textbf{the alignment of feature maps lead to the alignment of attention distributions}.

\section{Extra Experimental Settings}

For a fair comparison with original BERT, we follow the same pre-processing scheme as BERT, where we mask 15\% of all WordPiece \citep{kudo2018sentencepiece} tokens in each sequence at random and use next sentence prediction. Please note that MobileBERT can be potentially further improved by several training techniques recently introduced, such as span prediction \citep{joshi2019spanbert} or removing next sentence prediction objective \citep{liu2019roberta}. We leave it for future work.

In pre-training distillation, the hyperparameter $\alpha$ is used to balance the original masked language modeling loss and the distillation loss. Following \citep{kim2016sequence}, we set $\alpha$ to 0.5.

\begin{figure}[t]
	\centering
	\includegraphics[width=0.5\columnwidth]{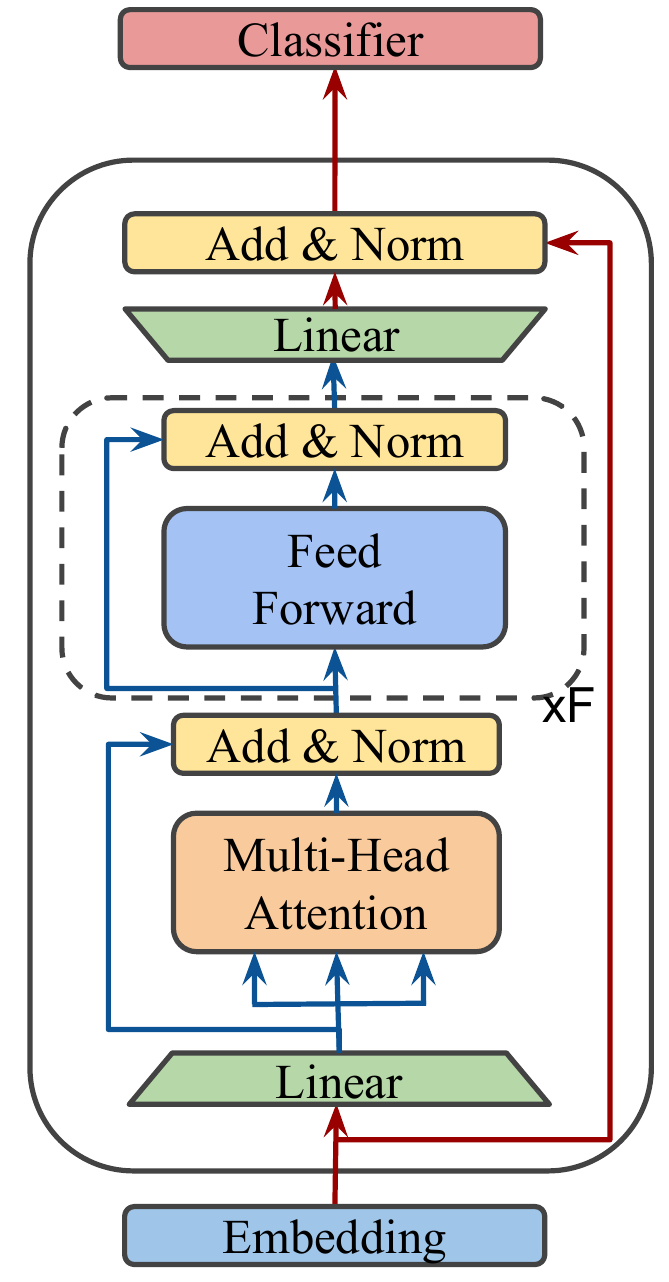}
	\caption{Illustration of $\text{MobileBERT}_\text{TINY}$.
	\textcolor{myred}{red lines denote {inter-block} flows} while \textcolor{myblue}{{blue lines} {intra-block} flows}.
	}
\label{fig:tiny}
\end{figure}

\section{Architecture of $\text{MobileBERT}_\text{TINY}$}

We use a lighter MHA structure for $\text{MobileBERT}_\text{TINY}$. As illustrated in Figure \ref{fig:tiny}, in stead of using hidden states from the inter-block feature maps as inputs to MHA, we use the reduced intra-block feature maps as key, query, and values in MHA for $\text{MobileBERT}_\text{TINY}$. This can effectively reduce the parameters in MHA modules, but might harm the model capacity.

\section{GLUE Dataset}\label{app:glue}
	
In this section, we provide a brief description of the tasks in the GLUE benchmark \citep{wang2018glue}.

\paragraph{CoLA} The Corpus of Linguistic Acceptability \citep{warstadt2018neural} is a collection of English acceptability judgments drawn from books and journal articles on linguistic theory. The task is to predict whether an example is a grammatical English sentence and is evaluated by Matthews correlation coefficient \citep{matthews1975comparison}.

\paragraph{SST-2} The Stanford Sentiment Treebank \citep{socher2013recursive} is a collection of sentences from movie reviews and human annotations of their sentiment. The task is to predict the sentiment of a given sentence and is evaluated by accuracy.

\paragraph{MRPC} The Microsoft Research Paraphrase Corpus \citep{dolan2005automatically} is a collection of sentence pairs automatically extracted from online news sources. They are labeled by human annotations for whether the sentences in the pair are semantically equivalent. The performance is evaluated by both accuracy and F1 score.

\paragraph{STS-B} The Semantic Textual Similarity Benchmark \citep{cer2017semeval} is a collection of sentence pairs drawn from news headlines, video and image captions, and natural language inference data. Each pair is human-annotated with a similarity score from 1 to 5. The task is to predict these scores and is evaluated by Pearson and Spearman correlation coefficients.

\paragraph{QQP} The Quora Question Pairs\footnote{\url{https://data.quora.com/First-Quora-Dataset-Release-Question-Pairs}} \citep{chen2018quora} dataset is a collection of question pairs from the community question-answering website Quora. The task is to determine whether a pair of questions are semantically equivalent and is evaluated by both accuracy and F1 score.

\paragraph{MNLI} The Multi-Genre Natural Language Inference Corpus \citep{williams2018broad} is a collection of sentence pairs with textual entailment annotations. Given a premise sentence and a hypothesis sentence, the task is to predict whether the premise entails the hypothesis (\emph{entailment
}), contradicts the hypothesis (\emph{contradiction}), or neither (\emph{neutral}) and is evaluated by accuracy on both \emph{matched} (in-domain) and \emph{mismatched} (cross-domain) sections of the test data.

\paragraph{QNLI} The Question-answering NLI dataset is converted from the Stanford Question Answering Dataset (SQuAD) \citep{rajpurkar2016squad}. The task is to determine whether the context sentence contains
the answer to the question and is evaluated by the test accuracy.

\paragraph{RTE} The Recognizing Textual Entailment (RTE) datasets come from a series of annual textual entailment challenges \citep{bentivogli2009fifth}. The task is to predict whether sentences in a sentence pair are entailment and is evaluated by accuracy.

\paragraph{WNLI} The Winograd Schema Challenge \citep{levesque2011winograd} is a reading comprehension task in which a system must read a sentence with a pronoun and select the referent of that pronoun from a list of choices. We follow \citet{devlin2018bert} to skip this task in our experiments, because few previous works do better than predicting the majority class for this task.

\end{document}